\documentclass[11pt,a4paper]{article}
\usepackage[hyperref]{acl2017}
\usepackage{times}
\usepackage{latexsym}
\usepackage{graphicx}
\usepackage{caption}
\usepackage{subcaption}
\usepackage{amsmath}
\usepackage{amssymb}
\usepackage{booktabs}
\usepackage{color}

\usepackage{url}

\aclfinalcopy 

\title{A Comparative Study of Word Embeddings \\for Reading Comprehension}

\author{Bhuwan Dhingra, Hanxiao Liu, Ruslan Salakhutdinov, William W. Cohen \\
  School of Computer Science \\
  Carnegie Mellon University, Pittsburgh, USA \\
  {\tt \{bdhingra, hanxiaol, rsalakhu, wcohen\}@cs.cmu.edu} 
}

\date{}

\begin{document}
\maketitle
\begin{abstract}
The focus of past machine learning research for Reading Comprehension tasks has been primarily on the design of novel deep learning architectures. Here we show that seemingly minor choices made on (1) the use of pre-trained word embeddings, and (2) the representation of out-of-vocabulary tokens at test time, can turn out to have a larger impact than architectural choices on the final performance. We systematically explore several options for these choices, and provide recommendations to researchers working in this area.
\end{abstract}

\section{Introduction}
Systems that can read documents and answer questions about their content are a key language technology. 
The field, which has been termed \textit{Reading Comprehension} (RC), has attracted a tremendous amount of interest in the last two years, primarily due to the introduction of large-scale annotated datasets, such as CNN \citep{hermann2015teaching}
and SQuAD \citep{rajpurkar2016squad}. 

Powerful statistical models, including deep learning models (also termed as \textit{readers}), have been proposed for RC, most of which employ the following recipe: (1) Tokens in the document and question are represented using word vectors obtained from a lookup table (either initialized randomly, or from a pre-trained source such as \textit{GloVe} \citep{pennington2014GloVe}). (2) A sequence model such as LSTM \citep{hochreiter1997long}, augmented with an attention mechanism \citep{bahdanau2014neural}, updates these vectors to produce contextual representations. (3) An output layer uses these contextual representations to locate the answer in the document. The focus so far in the literature has been on steps (2) and (3), and several novel architectures have been proposed (see Section~\ref{sec:methods}). 

\begin{figure}[t]
\centering
\includegraphics[width=0.7\linewidth]{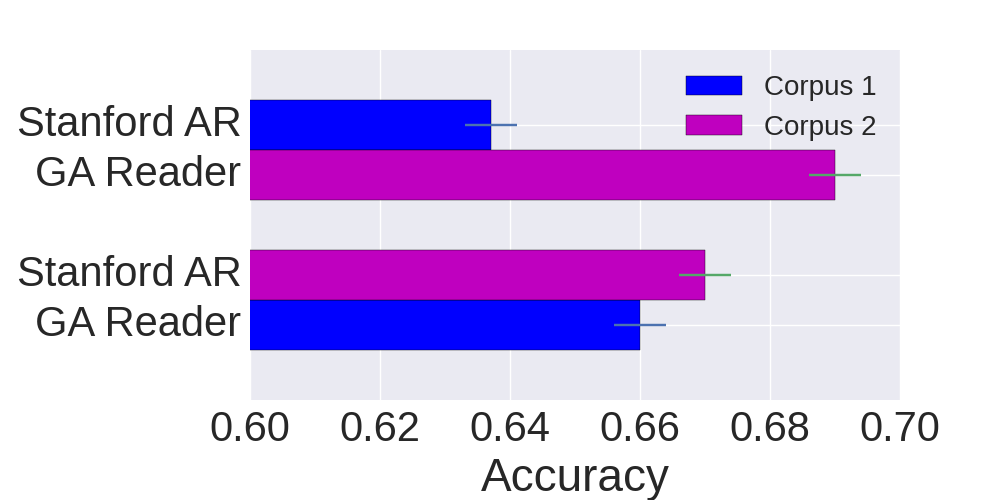}
\caption{\small Test set accuracies and std error on the Who-Did-What dataset for Stanford AR and GA Reader, trained after initializing with word vectors induced from different corpora. Without controlling for the initialization method, different conclusions may be drawn about which architecture is superior. Corpus 1: BookTest dataset \citep{bajgar2016embracing}, Corpus 2: Wikipedia + Gigaword.}
\label{fig:ga_sar}
\vspace{-0.6cm}
\end{figure}

In this work, we show that seemingly minor choices made in step (1), such as the use of pre-trained word embeddings and the handling of out-of-vocabulary tokens at test time, can lead to substantial differences in the final performance of the reader. These differences are usually much larger than the gains reported due to architectural improvements. As a concrete example, in Figure \ref{fig:ga_sar} we compare the performance of two RC models---Stanford Attentive Reader (AR) \citep{chen2016thorough} 
and Gated Attention (GA) Reader \citep{dhingra2016gated}---on the Who-Did-What dataset \citep{onishi2016did}, initialized with word embeddings trained on different corpora. Clearly, comparison between architectures is meaningful only under a controlled initialization method. 

To justify our claims,
we conduct a comprehensive set of experiments comparing the effect of utilizing embeddings pre-trained on several corpora. 
We experiment with RC datasets from different domains 
 using different architectures,
 and obtain consistent results across all settings. Based on our findings, we recommend the use of 
certain pre-trained GloVe vectors for initialization.
These consistently outperform other off-the-shelf embeddings such as \textit{word2vec}
\footnote{\scriptsize \url{https://code.google.com/archive/p/word2vec/}}
\citep{mikolov2013distributed}, as well as those pre-trained on the target corpus itself and, perhaps surprisingly, those trained on a large corpus from the same domain as the target dataset. 

Another important design choice 
is the handling of out-of-vocabulary (OOV) tokens at test time. 
A common approach (e.g. \citep{chen2016thorough,shen2016reasonet}) is to replace infrequent words during training with a special token \texttt{UNK}, and use this token to model the OOV words at the test phase. In reading comprehension, where the target answers are often rare words, we find that this approach leads to significantly worse performance in certain cases. A superior strategy is to assign each OOV token either a pre-trained, if available, or a random but unique vector at test time. We discuss and compare these two strategies, as well a mixed variant of the two, in 
Section~\ref{sec:oov}.

\section{Background}
\subsection{RC Datasets \& Models}
\label{sec:methods}
Many datasets aimed at measuring the performance of RC have been proposed \citep{nguyen2016ms,trischler2016newsqa}. For our purposes, we pick two of these benchmarks from different domains -- Who-Did-What (WDW) \citep{onishi2016did} constructed from news stories, and the Children's Book Test (CBT) \citep{hill2015goldilocks} constructed from children's books. For CBT we only consider the questions where the answer is a named entity (CBT-NE). 
Several RC models based on deep learning have been proposed~\cite{cui2016attention,munkhdalai2016neural,sordoni2016iterative,shen2016reasonet,kobayashi2016dynamic,henaff2016tracking,wang2016machine,wang2016multi,seo2016bidirectional,xiong2016dynamic,yu2016end}. For our experiments we pick two of these models: the simple, but competitive, Stanford AR, and the high-performing GA Reader.

\paragraph{Stanford AR:} The Stanford AR consists of single-layer Bidirectional GRU encoders for both the document and the query, followed by a bilinear attention operator for computing a weighted average representation of the document. 
The original model, which was developed for the anonymized CNN / Daily Mail datasets, used an output lookup table $W_a$ to select the answer. However, without anonymization the number of answer candidates can become very large. Hence, we instead select the answer from the document representation itself, followed by an \textit{attention sum} mechanism \citep{kadlec2016text}. This procedure is very similar to the one used in GA Reader, and is described in detail in Appendix \ref{app:sar}. 

\paragraph{GA Reader:} The GA Reader is a multi-hop architecture which updates the representation of document tokens through multiple bidirectional GRU layers \citep{cho2014learning}. At the output of each intermediate layer, the token representations are re-weighted by taking their element-wise product with an attention-weighted representation of the query. The outputs of the final layer are further matched with the query representation with an inner product to produce a distribution over the candidate answers in the document, and multiple mentions are aggregated using attention sum. 
We use the publicly available code\footnote{\scriptsize \url{https://github.com/bdhingra/ga-reader}}
with 
the default hyperparameter settings of~\cite{dhingra2016gated}, detailed in Appendix~\ref{app:hyperparams}. 

\subsection{Word Embedding methods}
\begin{table}[t]
\small
\centering
\begin{tabular}{@{}lllll@{}}
\toprule
\textbf{Emb.}     & \textbf{Corpus}                                                                            & \textbf{Domain}                                             & \textbf{\begin{tabular}[c]{@{}l@{}}Size\end{tabular}} & \textbf{\begin{tabular}[c]{@{}l@{}}Vocab\end{tabular}} \\ \midrule
OTS & \begin{tabular}[c]{@{}l@{}}Wiki + Gigaword / \\ GoogleNews\end{tabular} & \begin{tabular}[c]{@{}l@{}}Wiki / \\ News\end{tabular} & \begin{tabular}[c]{@{}l@{}}6B / \\ 100B\end{tabular}                        & \begin{tabular}[c]{@{}l@{}}400K / \\ 3M\end{tabular}          \\
WDW                 & Who-Did-What                                                                       & News                                                        & 50M                                                                         & 91K                                                           \\
BT                  & BookTest                                                                            & Fiction                                                     & 8B                                                                          & 1.2M                                                          \\
CBT                 & Children's BookTest                                                                & Fiction                                                     & 50M                                                                         & 48K                                                           \\ \bottomrule
\end{tabular}
\caption{\small Details of corpora used for training word embeddings. \textbf{OTS:} Off-The-Shelf embeddings provided with GloVe / word2vec. Corpus size is in \# of tokens.}
\label{tab:embeddings}
\vspace{-0.4cm}
\end{table}

The two most popular methods for inducing word embeddings from text corpora are \textit{GloVe} \citep{pennington2014GloVe} and \textit{word2vec} \citep{mikolov2013distributed}. These packages also provide off-the-shelf (OTS) embeddings trained on large corpora\footnote{The word2vec package contains embeddings for both capitalized and lowercase words. We convert all words to lowercase, and if a word has both lowercase and uppercase embeddings we use the lowercase version.}. While the GloVe package provides embeddings with varying sizes ($50$-$300$), word2vec only provides embeddings of size $300$. This is an important difference, which we discuss in detail later. We also train three additional embeddings, listed in Table~\ref{tab:embeddings}, including those trained on the target datasets themselves. 
In summary, we test with two in-domain corpora for WDW: one large (OTS) and one small (WDW), and two in-domain corpora for CBT: one large (BT) and one small (CBT).

When training embeddings, 
we set hyperparameters to their default values in the provided packages (see Appendix \ref{app:hyperparams} for details). This is by no means an optimal choice, in fact previous studies \citep{levy2015improving} have shown that hyperparameter choices may have a significant impact on downstream performance. However, training a single RC model can take anywhere from several hours to several days, and tuning hyperparameters for the embedding method on this downstream task is both infeasible and rarely done in practice. Instead, our objective is to provide guidelines to researchers using these methods out-of-the-box. 

\section{Experiments and Results}

\subsection{Comparison of Word Embeddings}
\begin{figure*}
\centering
\begin{subfigure}[t]{0.48\textwidth}
\includegraphics[width=\textwidth,height=3.5cm]{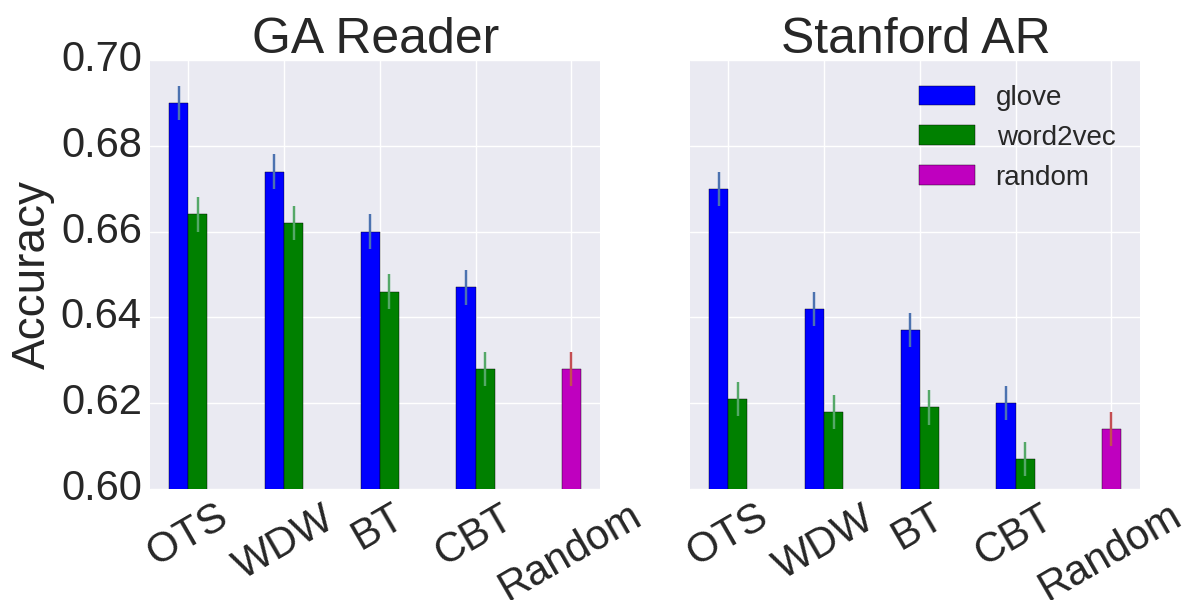}
\end{subfigure}
\begin{subfigure}[t]{0.48\textwidth}
\includegraphics[width=\textwidth,height=3.5cm]{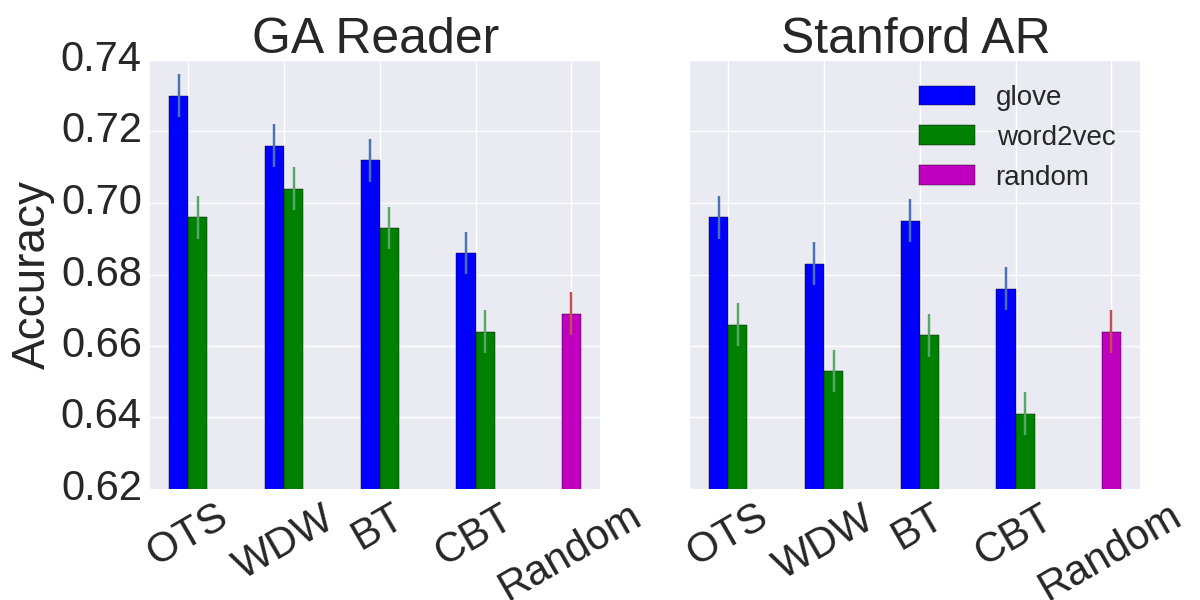}
\end{subfigure}
\vspace{-0.3cm}
\caption{\small Test set accuracy and std error for GA Reader and Stanford AR on WDW (left, middle-left) and CBT-NE (middle-right, right) when trained after initializing with pre-trained embeddings induced from different corpora (Table~\ref{tab:embeddings}), or randomly}
\label{fig:main}
\vspace{-0.5cm}
\end{figure*}

\begin{figure}
\centering
\begin{subfigure}[t]{0.49\linewidth}
\includegraphics[width=\textwidth,trim={2mm 0mm 10mm 8mm},clip]{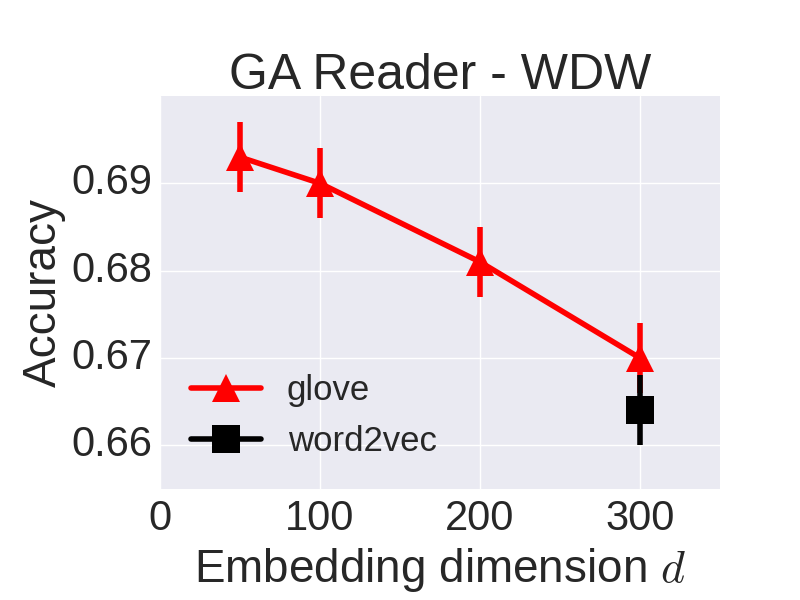}
\end{subfigure}
\begin{subfigure}[t]{0.49\linewidth}
\includegraphics[width=\textwidth,trim={2mm 0mm 10mm 2mm},clip]{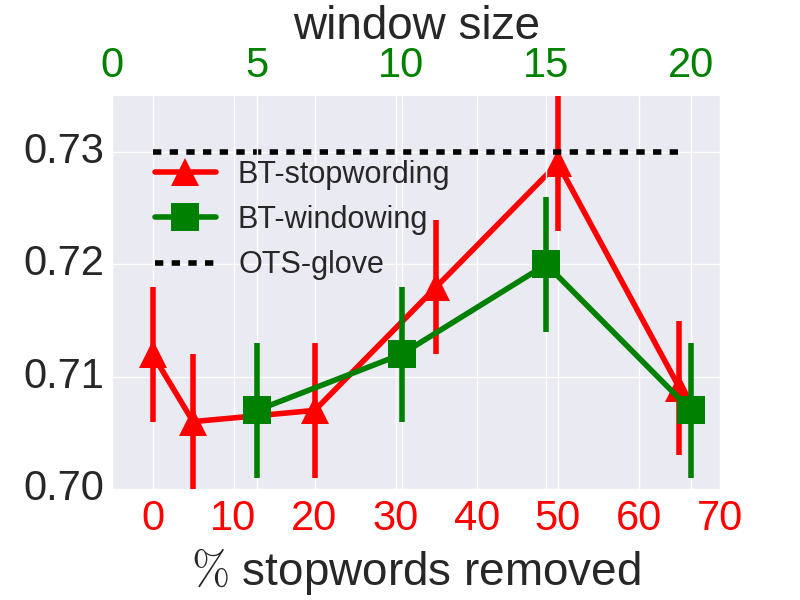}
\end{subfigure}
\vspace{-0.3in}
\caption{\small Test set accuracy and std error on \textbf{left:} WDW when initialized with off-the-shelf GloVe embeddings of different sizes, \textbf{right:} CBT-NE when initialized with embeddings trained on BT corpus after removing a fraction of stopwords ({\color{red} red}), or using different window sizes ({\color{green} green}).}
\label{fig:analysis}
\vspace{-0.4cm}
\end{figure}

We repeat each experiment twice with different random seeds and report the average test set accuracy across the two runs. Figure \ref{fig:main} shows a comparison of the RC performance for GA Reader and Stanford AR after initializing with various pre-trained embeddings, and also after initializing randomly. We see consistent results across the two datasets and and the two models.

The first observation is that using embeddings trained on the right corpora can improve anywhere from 3-6\% over random initialization. However, the corpus and method used for pre-training are important choices: for example word2vec embeddings trained on CBT perform worse than random. Also note that in every single case, GloVe embeddings outperform word2vec embeddings trained on the same corpora. It is difficult to claim that one method is better than the other, since previous studies \citep{levy2015improving} have shown that these methods are sensitive to hyperparameter tuning. However, if used out-of-the-box, GloVe seems to be the preferred method for pre-training. 

The single best performance is given by off-the-shelf GloVe embeddings ($d=100$) in each case, which outperform off-the-shelf word2vec embeddings ($d=300$). To understand if the difference comes from the differing dimension sizes, we plot the performance of GloVe embeddings as the dimension size is increased in Figure \ref{fig:analysis} (left). Performance drops as the embedding dimension size is increased (most likely due to over-fitting); however even at $d=300$, GloVe embeddings outperform word2vec embeddings. 

On both test datasets embeddings trained on formal domains, like news (OTS, WDW), perform at least as well as those trained on informal ones, like fiction (BT, CBT). This is surprising for CBT-NE dataset which is itself constructed from the informal domain of children's books. For example, WDW (50M tokens) does significantly better than CBT-NE (50M tokens) in $3$ out of the $4$ cases, and also significantly better than the much larger BT (8B tokens) in one setting (and comparably in other settings).
A key distinguishing feature between these two domains is the fraction of text composed of stopwords
\footnote{\scriptsize We use the list of stopwords available at \url{http://research.microsoft.com/en-us/um/redmond/projects/mctest/data/stopwords.txt}}
 -- WDW consists of 54\% stopwords while BT consists of 68\% stopwords. Both GloVe and word2vec induce word vectors by minimizing the Euclidean distance between vectors of frequently co-occurring words. Co-occurrence with stopwords, however, provides little meaningful information about the semantics of a particular word, and hence corpora with a high percentage of these may not produce high-quality vectors. This effect may be mitigated during pre-training by either, (1) removing a fraction of stopwords from the corpus, (2) increasing the window size for counting co-occuring words. Figure \ref{fig:analysis} (right) shows the effect of both these methods on downstream RC performance. There is an improvement as the fraction of stopwords decreases or the window size increases, upto a certain limit. In fact, with proper tuning, BT embeddings give roughly the same performance as OTS GloVe, emphasizing the importance of hyperparameter tuning when training word vectors. There is also evidence that stopword removal can be beneficial when training word vectors, however this needs further verification on other downstream tasks.

\subsection{Handling OOV tokens}
\label{sec:oov}

\begin{figure}
\centering
\includegraphics[width=\linewidth,height=3cm]{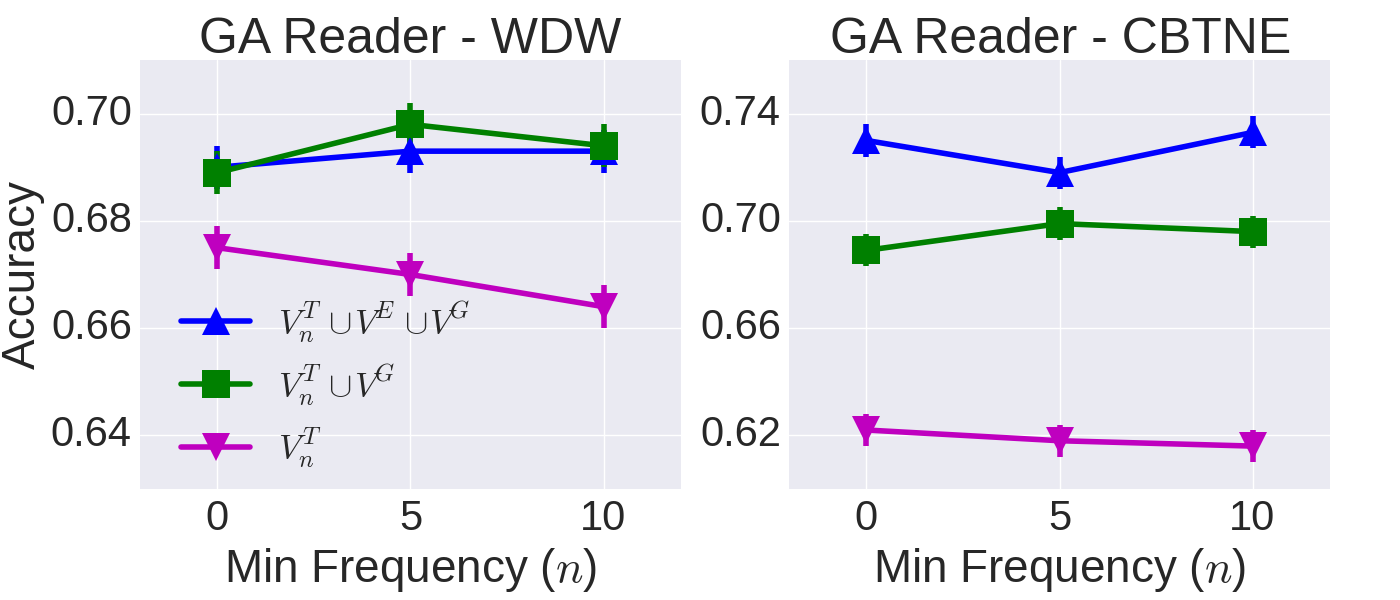}
\vspace{-0.3in}
\caption{\small Test set accuracy and std error for GA Reader on WDW (\textbf{left}) and CBT-NE (\textbf{right}) when trained after assigning different cuts of the vocabulary with word vectors. \textit{Min frequency} refers to the minimum count of a word type for it to be included in the vocabulary. }
\label{fig:oov}
\vspace{-0.5cm}
\end{figure}

In this section we study some common techniques for dealing with OOV tokens at test time. Based on the results from the previous section, we conduct this study using only the off-the-shelf GloVe pre-trained embeddings. Let the training, test and GloVe vocabularies be denoted by $V^T$, $V^E$ and $V^G$ respectively. Also define 
$V_n^T = \{t \in V^T : \#t > n\}$
where $\#t$ denotes the count of token $t$ in the training corpus. Before training a neural network for RC, the developer must first decide on the set of words $V$ which will be assigned word vectors. Any token outside $V$ is treated as an OOV token (denoted by \texttt{UNK}) and is assigned the same fixed vector. 

By far the most common technique in NLP literature (e.g. \citep{chen2016thorough,shen2016reasonet}) for constructing this vocabulary is to decide on a minimum frequency threshold $n$ (typically 5-10) and set $V=V_n^T$. Out of these, vectors for those which also appear in $V^G$ are initialized to their GloVe embeddings, and the rest are randomly initialized. Remaining tokens in $V^T$ and those in $V^E-V^T$ are all assigned the \texttt{UNK} vector, which is itself updated during training. This method ignores the fact that many of the words assigned as \texttt{UNK} may have already trained embeddings available in $V^G$. Hence, here we propose another strategy of constructing the vocabulary as $V=V_n^T \cup V^G$. 
Then at test time, any new token would be assigned its GloVe vector if it exists, or the vector for \texttt{UNK}. 
A third approach, used in \citep{dhingra2016gated}, is motivated by the fact that many of the RC models rely on computing fine-grained similarity between document and query tokens. Hence, instead of assigning all OOV tokens a common \texttt{UNK} vector, it might be better to assign them untrained but unique random vectors. This can be done by setting the vocabulary to $V=V^{T}_n \cup V^{E} \cup V^G$. Hence, at test time any new token will be assigned its GloVe vector if it exists, or a random vector. Note that for this approach access to $V^E$ at training time is not needed.

Figure \ref{fig:oov} shows a comparison of all three approaches with varying $n$ for the GA Reader on WDW and CBT-NE datasets. A gap of $3\%$ and $11\%$ between the best and worst setting for WDW and CBT-NE respectively clearly indicates the importance of using the correct setting. The commonly used method of setting $V=V^T_n$ is not a good choice for RC, and gets worse as $n$ is increased. It performs particularly poorly for the CBT-NE dataset, where $\sim 20\%$ of the test set answers do not appear in the training set (compared to only $\sim 1.5\%$ in WDW). The other two approaches perform comparably for WDW, but for CBT-NE assigning random vectors rather than \texttt{UNK} to OOV tokens gives better performance. This is also easily explained by looking at fraction of test set answers which do not occur in $V^T_0 \cup V^G$ -- it is $\sim 10\%$ for CBT-NE, and $<1\%$ for WDW. Since in general it is not possible to compute these fractions without access to the test set, we recommend setting $V=V^T_n \cup V^E \cup V^G$.

\section{Conclusions}
We have shown that the choice of pre-trained embeddings for initializing word vectors has a significant impact on the performance of neural models for reading comprehension. So does the method for handling OOV tokens at test time. We argue that different architectures can only be compared when these choices are controlled for. Based on our experiments, we recommend the use of off-the-shelf GloVe embeddings, and assigning pre-trained GloVe vectors, if available, or random but unique vectors to OOV tokens at test time.

\section*{Acknowledgments}
This work was funded by NSF under CCF1414030 and Google Research.

\bibliography{acl2017}

\begin{thebibliography}{}
\expandafter\ifx\csname natexlab\endcsname\relax\def\natexlab#1{#1}\fi

\bibitem[{Bahdanau et~al.(2014)Bahdanau, Cho, and Bengio}]{bahdanau2014neural}
Dzmitry Bahdanau, Kyunghyun Cho, and Yoshua Bengio. 2014.
\newblock Neural machine translation by jointly learning to align and
  translate.
\newblock {\em arXiv preprint arXiv:1409.0473\/} .

\bibitem[{Bajgar et~al.(2016)Bajgar, Kadlec, and
  Kleindienst}]{bajgar2016embracing}
Ondrej Bajgar, Rudolf Kadlec, and Jan Kleindienst. 2016.
\newblock Embracing data abundance: Booktest dataset for reading comprehension.
\newblock {\em arXiv preprint arXiv:1610.00956\/} .

\bibitem[{Chen et~al.(2016)Chen, Bolton, and Manning}]{chen2016thorough}
Danqi Chen, Jason Bolton, and Christopher~D Manning. 2016.
\newblock A thorough examination of the cnn/daily mail reading comprehension
  task.
\newblock {\em arXiv preprint arXiv:1606.02858\/} .

\bibitem[{Cho et~al.(2014)Cho, Van~Merri{\"e}nboer, Gulcehre, Bahdanau,
  Bougares, Schwenk, and Bengio}]{cho2014learning}
Kyunghyun Cho, Bart Van~Merri{\"e}nboer, Caglar Gulcehre, Dzmitry Bahdanau,
  Fethi Bougares, Holger Schwenk, and Yoshua Bengio. 2014.
\newblock Learning phrase representations using rnn encoder-decoder for
  statistical machine translation.
\newblock {\em arXiv preprint arXiv:1406.1078\/} .

\bibitem[{Cui et~al.(2016)Cui, Chen, Wei, Wang, Liu, and Hu}]{cui2016attention}
Yiming Cui, Zhipeng Chen, Si~Wei, Shijin Wang, Ting Liu, and Guoping Hu. 2016.
\newblock Attention-over-attention neural networks for reading comprehension.
\newblock {\em arXiv preprint arXiv:1607.04423\/} .

\bibitem[{Dhingra et~al.(2016)Dhingra, Liu, Cohen, and
  Salakhutdinov}]{dhingra2016gated}
Bhuwan Dhingra, Hanxiao Liu, William~W Cohen, and Ruslan Salakhutdinov. 2016.
\newblock Gated-attention readers for text comprehension.
\newblock {\em arXiv preprint arXiv:1606.01549\/} .

\bibitem[{Henaff et~al.(2016)Henaff, Weston, Szlam, Bordes, and
  LeCun}]{henaff2016tracking}
Mikael Henaff, Jason Weston, Arthur Szlam, Antoine Bordes, and Yann LeCun.
  2016.
\newblock Tracking the world state with recurrent entity networks.
\newblock {\em arXiv preprint arXiv:1612.03969\/} .

\bibitem[{Hermann et~al.(2015)Hermann, Kocisky, Grefenstette, Espeholt, Kay,
  Suleyman, and Blunsom}]{hermann2015teaching}
Karl~Moritz Hermann, Tomas Kocisky, Edward Grefenstette, Lasse Espeholt, Will
  Kay, Mustafa Suleyman, and Phil Blunsom. 2015.
\newblock Teaching machines to read and comprehend.
\newblock In {\em Advances in Neural Information Processing Systems\/}. pages
  1684--1692.

\bibitem[{Hill et~al.(2015)Hill, Bordes, Chopra, and
  Weston}]{hill2015goldilocks}
Felix Hill, Antoine Bordes, Sumit Chopra, and Jason Weston. 2015.
\newblock The goldilocks principle: Reading children's books with explicit
  memory representations.
\newblock {\em arXiv preprint arXiv:1511.02301\/} .

\bibitem[{Hochreiter and Schmidhuber(1997)}]{hochreiter1997long}
Sepp Hochreiter and J{\"u}rgen Schmidhuber. 1997.
\newblock Long short-term memory.
\newblock {\em Neural computation\/} 9(8):1735--1780.

\bibitem[{Kadlec et~al.(2016)Kadlec, Schmid, Bajgar, and
  Kleindienst}]{kadlec2016text}
Rudolf Kadlec, Martin Schmid, Ondrej Bajgar, and Jan Kleindienst. 2016.
\newblock Text understanding with the attention sum reader network.
\newblock {\em arXiv preprint arXiv:1603.01547\/} .

\bibitem[{Kingma and Ba(2014)}]{kingma2014adam}
Diederik Kingma and Jimmy Ba. 2014.
\newblock Adam: A method for stochastic optimization.
\newblock {\em arXiv preprint arXiv:1412.6980\/} .

\bibitem[{Kobayashi et~al.(2016)Kobayashi, Tian, Okazaki, and
  Inui}]{kobayashi2016dynamic}
Sosuke Kobayashi, Ran Tian, Naoaki Okazaki, and Kentaro Inui. 2016.
\newblock Dynamic entity representations with max-pooling improves machine
  reading.
\newblock In {\em NAACL-HLT\/}.

\bibitem[{Levy et~al.(2015)Levy, Goldberg, and Dagan}]{levy2015improving}
Omer Levy, Yoav Goldberg, and Ido Dagan. 2015.
\newblock Improving distributional similarity with lessons learned from word
  embeddings.
\newblock {\em Transactions of the Association for Computational Linguistics\/}
  3:211--225.

\bibitem[{Mikolov et~al.(2013)Mikolov, Sutskever, Chen, Corrado, and
  Dean}]{mikolov2013distributed}
Tomas Mikolov, Ilya Sutskever, Kai Chen, Greg~S Corrado, and Jeff Dean. 2013.
\newblock Distributed representations of words and phrases and their
  compositionality.
\newblock In {\em Advances in neural information processing systems\/}. pages
  3111--3119.

\bibitem[{Munkhdalai and Yu(2016)}]{munkhdalai2016neural}
Tsendsuren Munkhdalai and Hong Yu. 2016.
\newblock Neural semantic encoders.
\newblock {\em arXiv preprint arXiv:1607.04315\/} .

\bibitem[{Nguyen et~al.(2016)Nguyen, Rosenberg, Song, Gao, Tiwary, Majumder,
  and Deng}]{nguyen2016ms}
Tri Nguyen, Mir Rosenberg, Xia Song, Jianfeng Gao, Saurabh Tiwary, Rangan
  Majumder, and Li~Deng. 2016.
\newblock Ms marco: A human generated machine reading comprehension dataset.
\newblock {\em arXiv preprint arXiv:1611.09268\/} .

\bibitem[{Onishi et~al.(2016)Onishi, Wang, Bansal, Gimpel, and
  McAllester}]{onishi2016did}
Takeshi Onishi, Hai Wang, Mohit Bansal, Kevin Gimpel, and David McAllester.
  2016.
\newblock Who did what: A large-scale person-centered cloze dataset.
\newblock {\em EMNLP\/} .

\bibitem[{Pennington et~al.(2014)Pennington, Socher, and
  Manning}]{pennington2014GloVe}
Jeffrey Pennington, Richard Socher, and Christopher~D. Manning. 2014.
\newblock \href{http://www.aclweb.org/anthology/D14-1162}{Glove: Global vectors
  for word representation}.
\newblock In {\em Empirical Methods in Natural Language Processing (EMNLP)\/}.
  pages 1532--1543.
\newblock
  \href{http://www.aclweb.org/anthology/D14-1162}{http://www.aclweb.org/anthology/D14-1162}.

\bibitem[{Rajpurkar et~al.(2016)Rajpurkar, Zhang, Lopyrev, and
  Liang}]{rajpurkar2016squad}
Pranav Rajpurkar, Jian Zhang, Konstantin Lopyrev, and Percy Liang. 2016.
\newblock Squad: 100,000+ questions for machine comprehension of text.
\newblock {\em arXiv preprint arXiv:1606.05250\/} .

\bibitem[{Seo et~al.(2016)Seo, Kembhavi, Farhadi, and
  Hajishirzi}]{seo2016bidirectional}
Minjoon Seo, Aniruddha Kembhavi, Ali Farhadi, and Hannaneh Hajishirzi. 2016.
\newblock Bidirectional attention flow for machine comprehension.
\newblock {\em arXiv preprint arXiv:1611.01603\/} .

\bibitem[{Shen et~al.(2016)Shen, Huang, Gao, and Chen}]{shen2016reasonet}
Yelong Shen, Po-Sen Huang, Jianfeng Gao, and Weizhu Chen. 2016.
\newblock Reasonet: Learning to stop reading in machine comprehension.
\newblock {\em arXiv preprint arXiv:1609.05284\/} .

\bibitem[{Sordoni et~al.(2016)Sordoni, Bachman, and
  Bengio}]{sordoni2016iterative}
Alessandro Sordoni, Phillip Bachman, and Yoshua Bengio. 2016.
\newblock Iterative alternating neural attention for machine reading.
\newblock {\em arXiv preprint arXiv:1606.02245\/} .

\bibitem[{Trischler et~al.(2016)Trischler, Wang, Yuan, Harris, Sordoni,
  Bachman, and Suleman}]{trischler2016newsqa}
Adam Trischler, Tong Wang, Xingdi Yuan, Justin Harris, Alessandro Sordoni,
  Philip Bachman, and Kaheer Suleman. 2016.
\newblock Newsqa: A machine comprehension dataset.
\newblock {\em arXiv preprint arXiv:1611.09830\/} .

\bibitem[{Wang and Jiang(2016)}]{wang2016machine}
Shuohang Wang and Jing Jiang. 2016.
\newblock Machine comprehension using match-lstm and answer pointer.
\newblock {\em arXiv preprint arXiv:1608.07905\/} .

\bibitem[{Wang et~al.(2016)Wang, Mi, Hamza, and Florian}]{wang2016multi}
Zhiguo Wang, Haitao Mi, Wael Hamza, and Radu Florian. 2016.
\newblock Multi-perspective context matching for machine comprehension.
\newblock {\em arXiv preprint arXiv:1612.04211\/} .

\bibitem[{Xiong et~al.(2016)Xiong, Zhong, and Socher}]{xiong2016dynamic}
Caiming Xiong, Victor Zhong, and Richard Socher. 2016.
\newblock Dynamic coattention networks for question answering.
\newblock {\em arXiv preprint arXiv:1611.01604\/} .

\bibitem[{Yu et~al.(2016)Yu, Zhang, Hasan, Yu, Xiang, and Zhou}]{yu2016end}
Yang Yu, Wei Zhang, Kazi Hasan, Mo~Yu, Bing Xiang, and Bowen Zhou. 2016.
\newblock End-to-end answer chunk extraction and ranking for reading
  comprehension.
\newblock {\em arXiv preprint arXiv:1610.09996\/} .

\end{thebibliography}
\bibliographystyle{acl_natbib}

\appendix

\section{Answer Selection for Stanford AR}
\label{app:sar}
Using the notation from \citep{chen2016thorough}, let $\tilde{p}_1, \tilde{p}_2, \ldots, \tilde{p}_m$ be the contextual embeddings of the tokens in the document, and let $o$ be the attention-weighted document representation, then we compute the probability that token $i$ answers the question as:
\begin{equation}
P(a=d_i|d,q) = s_i = \text{softmax}(\tilde{p}_i^T o)
\end{equation}
The probability of a particular candidate $c \in \mathcal{C}$ as being the answer is then computed by aggregating the probabilities of all document tokens which appear in $c$ and renormalizing over the candidates:
\begin{equation}
    \Pr(c|d,q) \propto \sum_{i\in \mathbb{I}(c,d)} s_i
\end{equation}
where $\mathbb{I}(c,d)$ is the set of positions where a token in $c$ appears in the document $d$. 

\section{Hyperparameter Details}
\label{app:hyperparams}
For the WDW dataset we use hidden state size $d=128$ for the GRU and dropout with $p=0.3$. For CBT-NE dataset we use $d=128$ and dropout with $p=0.4$. The Stanford AR has only 1 layer as proposed in the original paper, while the GA Reader has $3$ layers. For Stanford AR dropout is applied to the input of the layer, and for GA Reader it is applied in between layers. Embeddings sizes for the word vectors were set to $d_w=100$ for all experiments, except those using off-the-shelf word2vec embeddings. To enable a fair comparison, we utilize the \textit{qe-comm} feature for Stanford AR, which was used in the implementation of GA Reader. Since our purpose is to study the effect of word vectors, we do not use character embeddings in our experiments.

We train the models using the ADAM \citep{kingma2014adam} optimizer with an initial learning rate of 0.0005, which is halved every epoch after the first 3 epochs. We track performance on the validation set, and select the model with the highest validation accuracy for testing. 

When training word vectors we retain the default settings provided with the GloVe and word2vec packages, with the only exception that window size was set to $10$ for both (to ensure consistency). For word2vec, we used skip-gram architecture with hierarchical softmax, and sub-sampled frequent words with a threshold $10^{-3}$ (see \citep{mikolov2013distributed} for details). For GloVe, we used $15$ iterations when training on the small WDW and CBT corpora, and $50$ iterations for the large BT corpus. In any corpus, words occurring less than $5$ times were filtered before training the word vectors. 

\end{document}